# A Directional Feature with Energy based Offline Signature Verification Network

Minal Tomar[*] and Pratibha Singh[*]

Department of Electrical & Electronics[*]

Malwa Institute of Technology

Indore (M.P.) 452016

(www.malwa-institute.com)

Department of Electronics and Instrumentation

Institute of Engineering and Technology

Devi Ahilya Vishwavidyalaya, Indore (M.P.) 452017

(www.iet.dauniv.ac.in)

minal2121@rediffmail.com[*]
prat_ibh_a@yahoo.com

**Abstract.** Signature used as a biometric is implemented in various systems as well as every signature signed by each person is distinct at the same time. So, it is very important to have a computerized signature verification system. In offline signature verification system dynamic features are not available obviously, but one can use a signature as an image and apply image processing techniques to make an effective offline signature verification system. Author proposes a intelligent network used directional feature and energy density both as inputs to the same network and classifies the signature. Neural network is used as a classifier for this system. The results are compared with both the very basic energy density method and a simple directional feature method of offline signature verification system and this proposed new network is found very effective as compared to the above two methods, specially for less number of training samples, which can be implemented practically.

## 1 Introduction

For human identification, the usage of biometrics is obvious and important in daily routine. Signature may be used as a biometrics as every signature is distinct. As signature has already used and accepted as an identification of the person who signed in so many systems, it is important to keenly observe the signature before having any conclusion about the signee. This gives rise to a computerized signature verification system. But, in many systems only image of a signature is available so offline signature verification seems more important than online signature verification system. The main problem is that a signature of a person may vary according to his/her mood, health etc. even the genuine signer may not copy his/her signature as it is. There will always a change observed. Then it seems somewhat difficult to distinguish between genuine signature and a forgery.

With reference to this, it becomes a subject of continuous research until a purely faithful system is found to rely on. This paper is like one of a small step forward towards achieving the goal of a developed system for signature verification. Before proceeding further it is important to know about different type of forgeries. Forgeries can be classified as [1]:

a)  Random forgery: It produces without any knowledge of signature shape or even the signer's name.
b)  Simple forgery: It is formed by knowing only the name of the signer's but without having any examples of signer's signature style.
c)  Skilled forgery: It is produced by looking at an original signature sample, attempting to imitate it as closely as possible.

This paper is developed with the mind set that random and simple forgeries is very easy to detect and refuse but the main issue is to caught the skilled forger. That's why; the complete paper is processed with skilled forgeries only and when forgery word is used then it should be noted here that it means the skilled forgery.

*Signature Characteristics*

Signature must be treated as an image because a person may use any symbol or letter or group of letters etc. (according to the choice) as a signature, so it may be possible that one can't separate the letters written or even can't understand what is written. So, the system used for analysis of signature must use the concepts of image processing. Most probable, it is possible that the signature of a signer varied for every sign but there must be some unique characteristic to identify the signature so that it can be used as biometrics. Some essential characteristics are listed below:

a) Invariant: It should be constant over a long period of time.
b) Singular: It must be unique to the individual.
c) Inimitable: It must be irreproducible by other means.
d) Reducible and comparable: It should be capable of being reduced to a format that is easy to handle and digitally comparable to others [2].

As signature also belongs to one of the parameters which satisfy the above characteristics so it may be use as a proof of identification of a person.

*Types of signature verification*

As stated above signature is nothing but a pattern of special arrangement of pixels (i.e. an image). Still, we can classify the signature verification system based on whether only the image of a signature is available or the individual is personally signing before the verification system. Based on this, broadly, signature verification methods can be classified as:

a) ON-Line or Dynamic Signature Verification system
b) OFF-Line or Static Signature Verification system

In off-line verification system, only static features are considered which rely purely on the signature's image but require less hardware. Whereas in case of on-line systems, dynamic feature are taken into consideration, which include the time when stylus is in and out of the contact with the paper, the total time taken to make signature and the position where the stylus is raised from and lowered onto the paper, number of break points, maximum/minimum pressure of stylus contact, speed etc [2], [3].

*Related Work*

The 1st Signature verification system developed in 1965 but use of neural network for verification has been started in decades of 90s. Reena Bajaj et. al worked on signature verification using multiple neural classifiers. The authors have used 3 different types of global features projection moments, upper and lower envelop based characteristics with feed forward neural net for verification purpose [12]. Emre et. al had used global, direction and grid features and SVM (Support Vector Machine) as classification method. Authors also compared the performance using SVM and ANN [13]. Fixed point arithmetic with different classifiers such as HMM (Hidden Markov Models), SVM and ED (Euclidean Distance Classifier) analysed. Fathi et. al have used conjugate gradient neural network with string distance (SD) as local feature [16]. Enhanced modified directional feature and neural based classifiers have been used by Armand et. al [9]. Siddiqi et. al extracted chain code based features used ROC (Receiver Operating Characteristic) curve for classification [8]. Image registration, image fusion & DWT (Discrete Wavelet Transforms) are also used with Euclidean Distance Method for identification and verification of signatures. H. N. Prakash & D. S. Guru presented a new approach based on score level fusion for off-line signature verification [6]. Authors have used distance & orientation features to form bi-interval valued symbolic representation of signature. They compare the extracted features for classification. Author also proposed both the simple energy density method [17] and chain code method [18] for offline signature verification system in 2010.

This paper proposes a system for off-line signature verification using a directional feature mixed with energy density features extracted locally and Feed Forward Back Propagation Neural Network used as a classifier. Aspect ratio is also included as a global feature in energy density method.

## 2 Proposed Approach

There are three approaches used in this paper for feature extraction. First one is 'The Energy Density method' and the second is 'The Directional Feature Method'. Author used thinning algorithm because he database extracted from the contour obviously will acquire more memory as compared to the thinned image. Also a comparative statement between the simplest energy density method, Directional Feature method and obviously the proposed directional feature with energy density method is developed so that it may be clear that is it worth to improve the accuracy on the cost of memory and time?

International Journal on Soft Computing ( IJSC ), Vol.2, No.1, February 2011

## 2.1 Energy Density [17]

In this method, two features are used for training. Aspect ratio is used as a global feature and energy density is used as local feature. Aspect ratio is the ratio of Height (maximum vertical distance) to length (maximum horizontal distance) of the signature. We have calculated it after skew removal. Energy density is defined as the total energy present in each segment. We have done 100 segments of each signature and energy density is obtained by counting the total number of 1s in each segment (i.e. Total White Pixels). Thus, we have a feature vector of size 101X1 for energy density method as final database. This final database is fed to the neural network to perform the desired function i.e. training or classification.

## 2.2 Chain-Code [18]: As a directional Feature

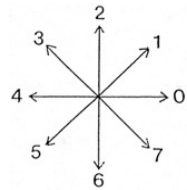

**Fig. 1.** 8 Connectivity of a Pixel

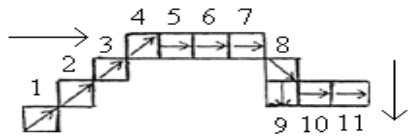

**Fig. 2.** Direction Changes in a Part of a Signature

Chain-code is based on the direction of the connecting pixels. Each pixel is observed for next connected pixel and their direction changes mapped by giving a numerical value to every possible direction. There are generally 8 connectivity is taken into consideration as shown in the Fig. 1. But in this paper we have used 4 connectivity i.e. 0, 1, 2 & 3. As another 4 directions i.e. 4, 5, 6 & 7 are simply the negation of 0, 1, 2 & 3 directions. To obtain chain-code top left corner is considered as origin and scanning is done left to right, top to bottom (refer Fig. 2). Each pixel has observed separately and direction vector for that pixel is noted down. This process is carried out until the last pixel has scanned. Now, the frequency of occurrence in a particular direction is calculated for each segment and the histogram for each segment is used it to train the neural network.

## 3 Implementation and Verification

Implementation of the proposed approach is basically divided in two parts i.e. Training and Classification. Training belongs to preparing and training of neural network for doing the classification work with an optimum accuracy. The proposed signature verification system takes an image of the signature as input and verifies whether the input image matches with the genuine training signature image available in the database or not. The system can be broadly categorized on the basis of method used for pre-processing and feature extraction from the image database and final input given to the neural network.

Raw Signature Database is gathered from 10 people and 110 Genuine & 110 Forgeries from each individual is collected (i.e. 2200 Signature Samples) and digitized using scanner, 300 dpi resolution. The first step for pre-processing is Binarization. It is used to produce binary image i.e. to convert colored (if any) image in black & white (i.e. in 0 or 1) format. In this paper global threshold is used for this purpose. Noise is filtered out using median filter. Thinning is done after noise filtering. Morphological operation is applied (in MATLAB) to performs the desired thinning. Rotation of signature patterns by a non predictive angle is one of the major difficulties. In this paper simply the concept of trigonometry is used for skew removal. Fig. 4 shows the effect of different stages of pre-processing. The next step of pre-processing is to extract the signature only from the whole image, by removing the image of extra paper remained. After signature extraction again resizing is carried out, and then segmentation process is completed to extract the local features of the signature (i.e. energy density or directional feature of each segment according to the method under test). Author has used 100 segment of each signature sample for further processing. In energy density method author has also used aspect ratio as a global feature to improve the performance of the system. Aspect ratio is extracted just before the resizing and segmentation.

In proposed directional feature with energy method for offline signature verification system author simply merge the above two methods altogether to observe the effect of the merging. Energy, aspect ration and directional feature of a thinned and binarized signature image is calculated and fed as an input to a feed forward back propagation neural network for training and classification again this preprocessing is done in the same manner. Fig. 3 shows the effect of some of the preprocessing steps. (a) shows the original digitized signature image under consideration. Then from (b) to (e) result of binarization, denoising, thining and skew removal is shown respectively. (f) describes the result of segmentation process for which 100 segments is made for each signature and further processing is done with each segments found after this step. After complete preprocessing both the energy density and directional feature is computed and fed to an artificially intelligent network for training and classification.

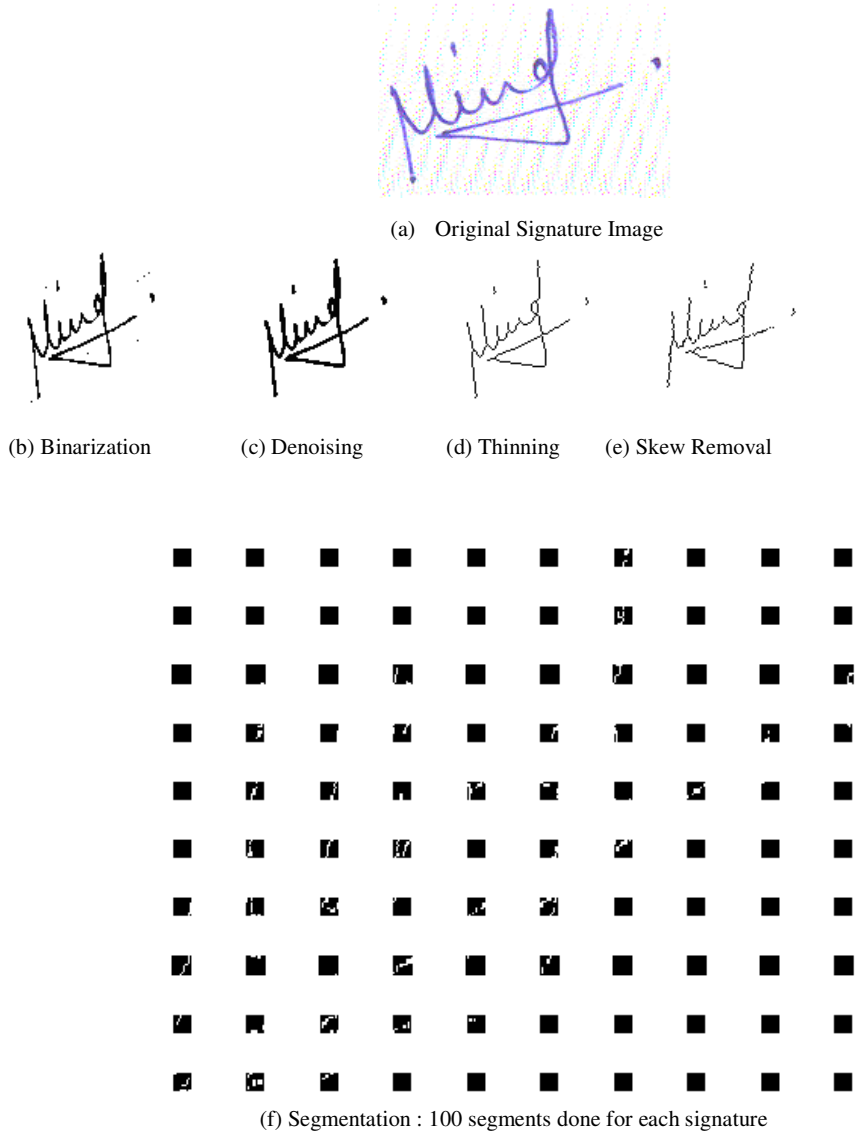

(a) Original Signature Image

(b) Binarization     (c) Denoising     (d) Thinning     (e) Skew Removal

(f) Segmentation : 100 segments done for each signature

**Fig. 3. (a) to (f)** Output of Different Stages of Pre-processing

**3.1 Proposed Architecture of Neural Network**

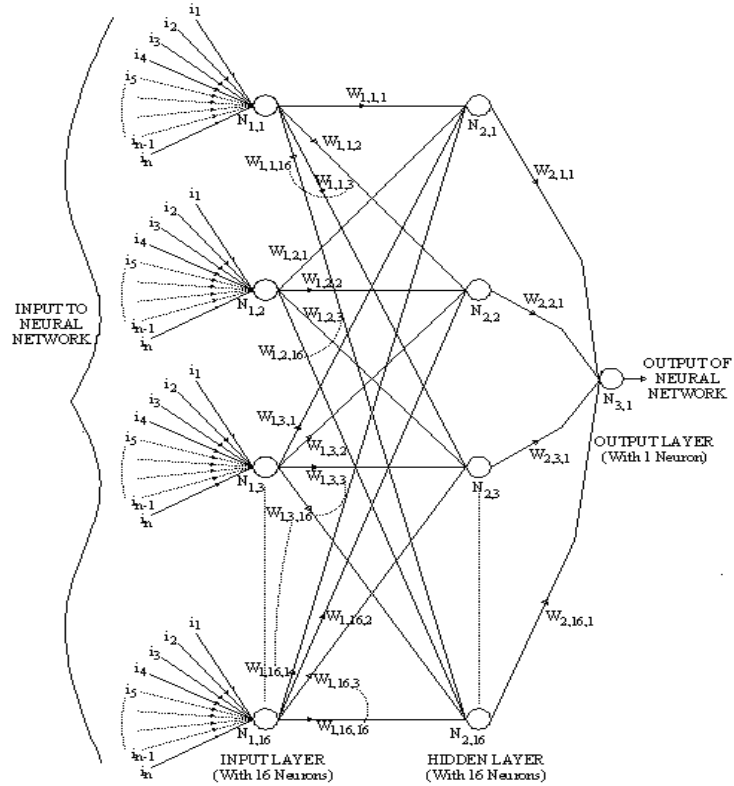

**Fig. 4.** Proposed Architecture of Neural Network

$N_{m,n}$ shows Neuron where, m = Layer Number & n = Neuron Number
$W_{i,j,k}$ represents Weight
where, i=Layer No.., j=Neuron No., k=Output No.of the particular Neuron
A Directional Feature with Energy based Offline Signature Verification Network

The proposed ANN scheme uses a multi layer feed forward network employing a back propagation learning algorithm with 16 Neurons in input layer and 1 Neuron in output layer. One hidden layer is present with 16 Neurons. The transfer function used for all the three layers are Hyperbolic Tangent Sigmoid (tansig). The proposed architecture of Neural Network is shown in Fig. 4. Here, default value of bias is chosen. Total 501 inputs i.e. 100 for energy of each segment, 400 as direction feature and 1 shows aspect ratio is given to this neural network.

### 3.2    Training of Neural Network

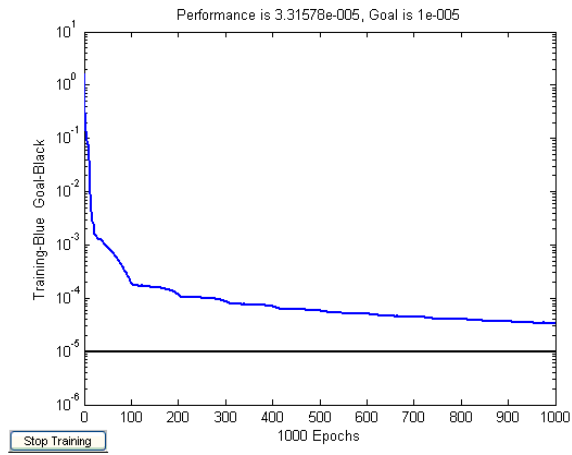

**Fig. 5.** Training of Neural Network

In this paper a supervised training is used. For this purpose 'Variable Learning Rate (traingdx) is used as a training function. Gradient descent with momentum weight and bias learning function (learngdm) is used by default. Mean square error (mse) is network performance function. It measures the network's performance according to the mean of squared errors. All other parameters of the neural network are set as by default including the initial weights. It is cleared by the graph in Fig. 5 that as the number of epochs are increasing during training then accordingly the mean square error is reducing. The black horizontal line is showing the reference (set at the time of preparing of NN) and the blue is denoting the reducing mse i.e. actual mse. The goal of training is to reduce mse till the extent that it will meet the reference. The training will stop when either the set goal has achieved or the maximum number of epochs reached.

## 4   Results

For comparison and performance evaluation of the proposed methodology we have used equal number of genuine and forgery samples for training and 100 numbers (50 Genuine + 50 Forgeries) are used for classification. Author increased the training signature samples from 10 (5 Genuine + 5 Forgeries) to 100 (50 Genuine + 50 Forgeries) and observe the effect of increasing training samples for all of the methods. For performance evaluation author used some common parameters like time required for training, accuracy, FAR (False Acceptance Ratio) i.e. the percentage of forgeries accepted as genuine & FRR (False Rejection Ratio) i.e. the percentage of rejecting the genuine signatures. A comparison for all the three methods has done on the basis of above mentioned parameters. Table I, II, III & IV along with graphs respectively of Fig. 6 elaborate the comparison of all the methods.

**TABLE I:** Comparison on the basis of Time Required for Training

| S. No. | No. of Taining Samples (50% Genuine + 50% Forgery) | Result | | |
|---|---|---|---|---|
| | | Elapsed Time (in Sec.) (Energy Density Method) | Elapsed Time (in Sec.) (Directional feature only) | Elapsed Time (in Sec.) (Directional feature with Energy  Method) |
| 1 | 10 | 6.156 | 7.375 | 7.672 |
| 2 | 20 | 6.453 | 7.89 | 8.718 |
| 3 | 30 | 6.593 | 8.39 | 9.141 |
| 4 | 40 | 7.312 | 9.282 | 9.594 |
| 5 | 50 | 6.625 | 10.718 | 11.64 |
| 6 | 60 | 7.671 | 8.172 | 8.5 |
| 7 | 70 | 7.797 | 8.281 | 8.5 |
| 8 | 80 | 6.313 | 8.594 | 11.688 |
| 9 | 90 | 6.704 | 9.188 | 9.531 |
| 10 | 100 | 6.969 | 9.125 | 9.532 |

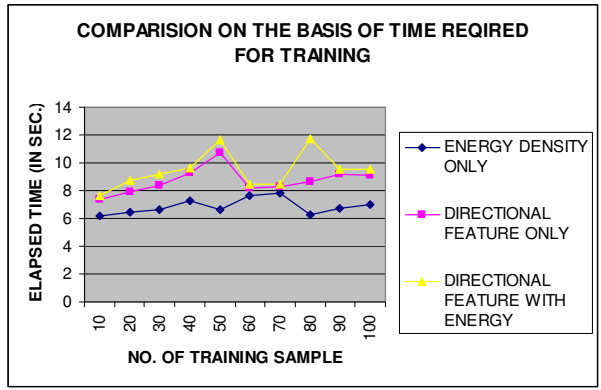

**(a)** Comparison for the time required for Training of NN

TABLE II: Comparison on the basis of Accuracy

| S. No. | No. of Taining Samples (50% Genuine + 50% Forgery) | Result | | |
|---|---|---|---|---|
| | | Accuracy (in%) (Energy Density Method) | Accuracy (in%) (Directional feature only) | Accuracy (in%) (Directional feature with Energy Method) |
| 1 | 10 | 57 | 70 | 71 |
| 2 | 20 | 77 | 74 | 87 |
| 3 | 30 | 68 | 78 | 84 |
| 4 | 40 | 70 | 70 | 84 |
| 5 | 50 | 67 | 75 | 83 |
| 6 | 60 | 80 | 83 | 86 |
| 7 | 70 | 88 | 89 | 93 |
| 8 | 80 | 96 | 97 | 97 |
| 9 | 90 | 94 | 96 | 98 |
| 10 | 100 | 96 | 97 | 100 |

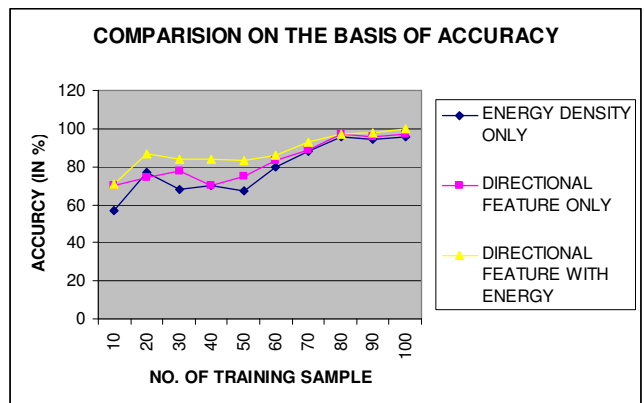

**(b)** Comparison for Accuracy

TABLE III: Comparison on the basis of FAR

| S. No. | No. of Taining Samples (50% Genuine + 50% Forgery) | Result | | |
|---|---|---|---|---|
| | | FAR (in%) (Energy Density Method) | FAR (in%) (Directional feature only) | FAR (in%) (Directional feature with Energy Method) |
| 1 | 10 | 42 | 40 | 28 |
| 2 | 20 | 42 | 40 | 12 |
| 3 | 30 | 42 | 30 | 10 |
| 4 | 40 | 48 | 40 | 6 |
| 5 | 50 | 46 | 20 | 8 |
| 6 | 60 | 22 | 6 | 4 |
| 7 | 70 | 2 | 0 | 6 |
| 8 | 80 | 6 | 0 | 0 |
| 9 | 90 | 10 | 0 | 0 |
| 10 | 100 | 6 | 4 | 0 |

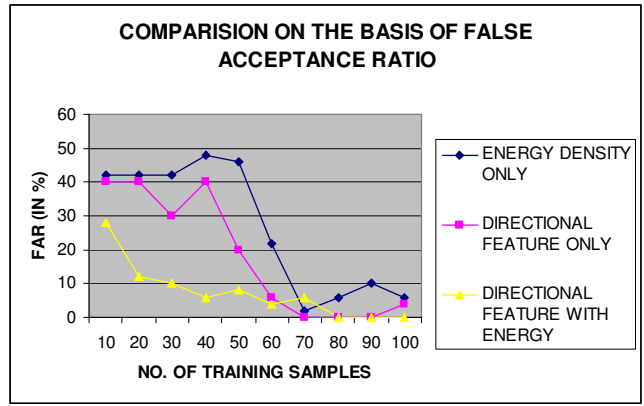

**(c)** Comparison for FAR

TABLE IV: Comparison on the basis of FRR

| S. No. | No. of Taining Samples (50% Genuine + 50% Forgery) | Result | | |
|---|---|---|---|---|
| | | FRR (in%) (Energy Density Method) | FRR (in%) (Directional feature only) | FRR (in%) (Directional feature with Energy Method) |
| 1 | 10 | 44 | 20 | 30 |
| 2 | 20 | 4 | 12 | 14 |
| 3 | 30 | 22 | 14 | 22 |
| 4 | 40 | 12 | 20 | 26 |
| 5 | 50 | 20 | 30 | 26 |
| 6 | 60 | 18 | 28 | 24 |
| 7 | 70 | 22 | 22 | 8 |
| 8 | 80 | 2 | 6 | 6 |
| 9 | 90 | 2 | 8 | 4 |
| 10 | 100 | 2 | 2 | 0 |

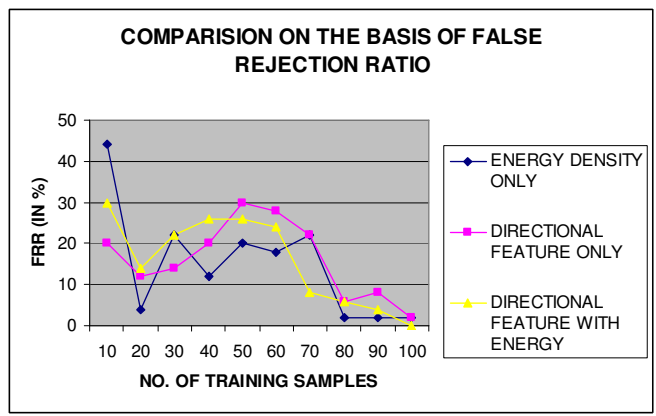

**(d)** Comparison for FRR

**Fig. 6 (a), (b), (c), (d).** shows the Comparison between Energy Density, Directional Feature Method and the proposed Directional feature with energy method

## 5 Conclusion

Author has compared the performance of a simplest known method (energy density method) [17], the directional feature Method [18] and the proposed directional feature with energy method on the basis time required for training, accuracy, False Acceptance Ratio & False Rejection Ratio. After many experiments and observations author comes to the conclusion that if the samples available for training are limited then directional feature method is better then energy density but at the same time directional feature merge with the energy method shows the best of all three methods. Although time required for training is slightly greater in case of this proposed method but the accuracy and FAR are satisfactory, and that extra time can be bearable. If one has a huge database for both Genuine & Forgeries to use for training then energy density method can be used but overall performance of this proposed method is far better then previous two. It is also observed that accuracy of energy density method is increasing rapidly as training sample increases but other two methods shows almost considerable results for all the training sample sets. Author has also observed that the results for FRR are varying randomly for all the cases. It gives the reason for future studies again.

The proposed directional feature with energy method for offline signature verification method gives far better results then the other two methods and can be implemented as easily as the previous one.

**Acknowledgments.** We are very thankful to the HOD (Electronics & Instrumentation Engineering), IET, DAVV, Indore and the Director, Malwa Institute of Technology, Indore for their constant support and motivation.

International Journal on Soft Computing ( IJSC ), Vol.2, No.1, February 2011